\documentclass[10pt,twocolumn,letterpaper]{article}

\usepackage[pagenumbers]{cvpr} 

\definecolor{cvprblue}{rgb}{0.21,0.49,0.74}
\usepackage[pagebackref,breaklinks,colorlinks,allcolors=cvprblue]{hyperref}

\title{Context-Aware Feature-Fusion for Co-occurring Object Detection in Autonomous Driving}

\author{
 Binay Kumar Singh \qquad Niels Da Vitoria Lobo \\
 Department of Computer Science \\
 University of Central Florida \\
 Orlando, USA \\
 {\tt\small \{binay.singh, niels.davitorialobo\}@ucf.edu}
}

\begin{document}


\maketitle

\begin{abstract} 
\label{sec:abstract} 
Object detection in autonomous driving requires precise localization and an inherent understanding of the relational context between co-occurring objects. In extremely complex heterogeneous environments rare classes, small-scale objects, and frequently appearing objects are difficult for standard object detection frameworks to handle. In this paper, we propose a novel framework called Context-Centric Feature Fusion (CCFF), which utilizes two attention-based modules, Local Context Fusion Module (LCFM) uses the RoI-to-RoI self-attention mechanism to resolve spatial interactions, mainly considering small and partially obscured objects, while Global Context Attention Module (GCAM) converts the co-occurrence of objects priors by pooling top-K RoI features into a global context attention token, avoiding the computational overhead of pixel-level global pooling. This fusion of local and object-centric global features yields contextualized embeddings that enhance classification results and co-occurring objects detection. Our method is evaluated on two datasets, Cityscapes and BDD100K which demonstrate significant improvement on relational consistency, achieving a Category-level Consistency Strategy (CCS) of 0.973 and 0.969, respectively. Furthermore, our approach produces substantial gains in small object detection ($AP_S$: 14.1\%) and successfully recovers rare classes such as “Train” that are typically lost in large distributions. Our efficiency report shows that the framework processes images in real time with a 0.2 FPS overhead. The code is available at 
\url{https://github.com/BinayKSingh/CCFF}.

\end{abstract}

\section{Introduction}
\label{sec:intro}

A key element of autonomous driving is object detection, which enables the model to perform subsequent operations \cite{Guo2025} such as tracking, motion forecasting, planning, and decision-making. Modern detectors, such as \cite{carion2020end} have achieved bottleneck performance in benchmark datasets, however, real driving scenes remain challenging due to frequent occlusions, cluttered intersections, diverse categories, and visually ambiguous instances (e.g., partially visible pedestrians or overlapping vehicles)  \cite{Li2025}. In such scenarios, accurate detection depends not only on the local appearance of an object but also on the context \cite{zhang2023dynamic} in which it appears. In modern object detectors, small and low-frequency objects \cite{hua2025elft}, and objects that appear under heavy occlusion or in dense traffic scene, remain a challenging problem.

Inevitably, the structure of roads and contexts \cite{mottaghi2014role} in which autonomous driving operates is very much fixed in real-life environments. In many road scenes, intersections and occluded objects co-occur, pedestrians are commonly seen across sidewalks and crosswalks, and dense traffic patterns often indicate particular object distributions. These co-occurrence features presented here provide a strong prior guideline that can determine these complex environments. However, many traditional detection pipelines still primarily rely on region-level appearance features extracted around each proposal, which limits their ability to exploit object-object relationships and global scene cues. 

Based on these rationale, we propose a model that understands how objects appear together in a scene. Our main contribution in this research work is as follows.

\begin{itemize}
    \item  We present a novel \textbf{Context-Centric Feature Fusion (CCFF)} framework that combines local context fusion module and object-centric global context module. This helps the model to learn about local and global contextual information required to enhance co-occurring object detection performance. 
    
    \item  First, we introduce \textbf{Local Context Fusion Module (LCFM)} by using RoI-to-RoI attention to model nearby object interactions that commonly occurs in crowded traffic scenes.
    
    \item Second, our object-centric \textbf{Global Context Attention Module (GCAM)} that integrates global context through attention pooling over RoI features, explicitly encode object co-occurrence priors in high-level environmental cues relevant to autonomous driving. Finally, by fusing on original RoI appearance features with locally and globally enhanced context features we produce \textbf{context-enriched embeddings} for classification and contextual priors.
    
    \item We evaluate our proposed work on Cityscapes and BDD100K datasets, demonstrating improved detection performance and co-occurring objects identification, specifically for partially occluded objects.

\end{itemize}

\section{Related Work}
\label{sec:related work}

\subsection{Attention-Based Contextual Reasoning}
For visual recognition tasks, transformer-based attention mechanisms have become an interesting tool for modeling long-range dependencies. By incorporating global contextual information, transformer-based architectures and attention-augmented convolutional networks \cite{wang2018nonlocal} have proven to perform well. Attention has been used at the feature-map level in object detection to capture semantic and spatial context \cite{hu2018relationnet, cao2019gcnet}. Despite their effectiveness, these techniques lack explicit reasoning over instance-level representations generated by region-based detectors and usually work with dense backbone features.

\subsection{Instance-Level and Relational Context}
To model object-object instance-level visual relationships, several studies have been investigated in the literature.  Attention mechanisms are introduced to capture pairwise interactions between detected objects by methods such as  Relation Networks \cite{hu2018relationnet} and object relation modules. Understandably, techniques mentioned here improve object localization and recognition by benefiting from instance-level context. The majority of relational reasoning frameworks, however, are only suitable as lightweight plug-and-play modules within conventional two-stage detectors because they are either made for relationship prediction tasks or require significant architectural changes.

\subsection{Global Context in Two-Stage Detectors}
To supplement local appearance features with scene-level cues, object detectors have also integrated global context. Previous research usually uses pooling over whole feature maps \cite{wang2018nonlocal, cao2019gcnet} or other context branches \cite{Bell2016ION} to aggregate global information. Recently introduced transformer-based detectors use self-attention across tokens to implicitly encode global context \cite{carion2020end, zhu2021deformable}. However, the methods mentioned here frequently have higher computational costs or architectural complexity and do not explicitly model object co-occurrence priors at the ROI level.

\subsection{Query-Based Vision Transformers vs. Region Priors}
Recent paradigms in perception have shifted toward query-based transformer detectors, such as Deformable DETR \cite{carion2020end} and DINO \cite{zhu2021deformable}. While these networks implicitly capture small environment constraints layer-by-layer via broad self-attention maps, they suffer from prolonged training convergence, massive computational overhead, and a high deployment overload. For highly active autonomous systems, extracting explicit, instance-level spatial co-occurrences directly within structured region-of-interest (RoI) bounds offers a much more deployment-viable alternative. Our CCFF framework bridges this gap by introducing local-global relational reasoning strictly inside the local prediction head, resolving complex spatial occlusions without the underlying model complexity or convergence overhead typical for pure vision transformers.

\begin{figure}[t]
  \centering
  \includegraphics[width=\linewidth]{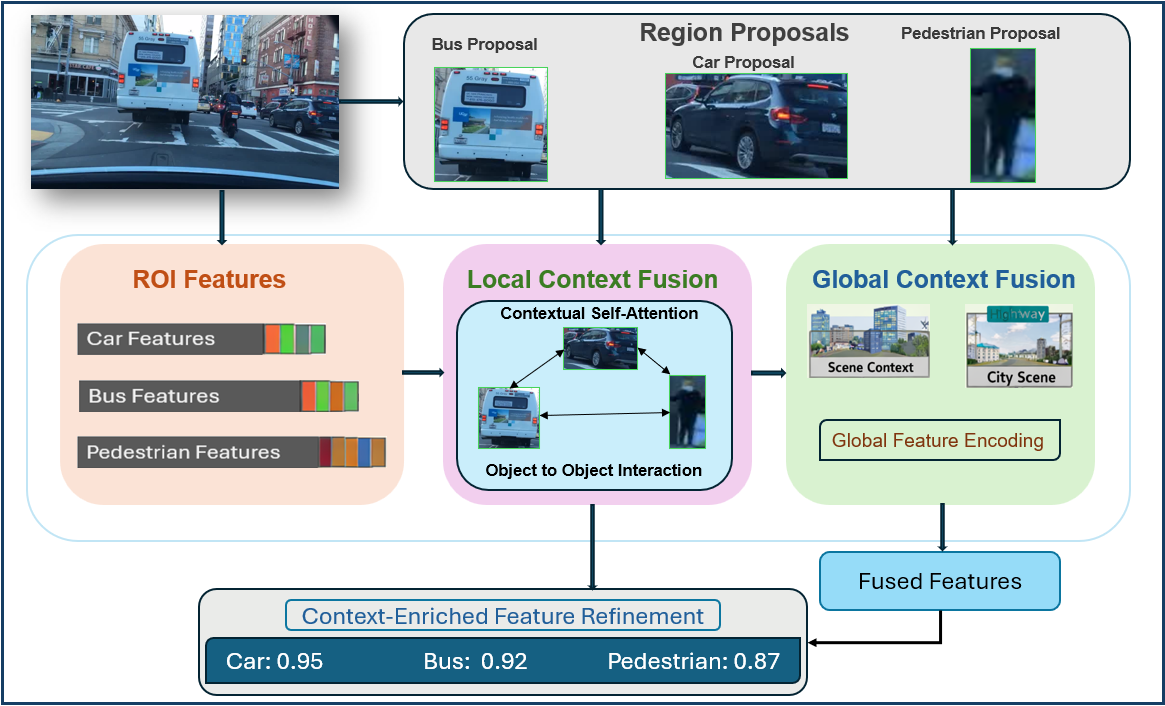}

\caption{\textbf{Detailed schematic flow of the proposed Context-Centric Feature Fusion (CCFF) architecture.} Region proposals from the FPN are enhanced using parallel dual-stream contextual reasoning channels. Our (1) the Local Context Fusion Module (LCFM), handles localized spatial object interactions using regional self-attention, (2) the Global Context Attention Module (GCAM), maps global environmental priors by executing attention pooling over the top-$K$ confident proposals biased by their normalized bounding coordinates. Then fused, context-enriched features pass directly to classification and regression heads to resolve ambiguities without relying on sequential tracking streams.}

\label{fig:res}

\end{figure}

\section{The Proposed Method}
\label{sec:method}

\subsection{Overview}
\label{sec:overview}
Our proposed framework Context-Centric Feature Fusion (CCFF) is presented in Fig. ~\ref{fig:res}. Given an input image $I \in \mathbb{R}^{H \times W \times 3}$, we build our framework with a Detectron2 backbone and Feature Pyramid Network (FPN) that extracts multi-scale feature maps $\{F_\ell\}_{\ell=1}^{L}$, and a Region Proposal Network (RPN) produces a set of region proposals $\mathcal{P}=\{p_i\}_{i=1}^{N}$. Each proposal is mapped to a fixed-dimensional RoI feature via RoIAlign:
\begin{equation}
\mathbf{r}_i = \mathrm{RoIAlign}(F, p_i) \in \mathbb{R}^{d},
\label{eq:roi}
\end{equation}
where $F$ denotes the respective pyramid feature level(s) selected for the proposal.

While conventional RoI heads process each $\mathbf{r}_i$ largely independently, our goal is to enrich RoI representations with contextual cues that are critical in autonomous driving scenes. To this end, we introduce two complementary context modules: (i) \emph{local context modeling} via RoI-to-RoI attention and (ii) \emph{object-centric global context modeling} via attention pooling over RoIs to encode co-occurrence priors at the scene level \cite{hua2025elft}. The resulting features are fused and passed to standard classification and bounding box regression heads. 

\subsection{Local Context Fusion Modeling via RoI-to-RoI Attention}
\label{sec:local}
In fixed-range contextual features there is a limitation, addressed recently in Dynamic Context Exploration (DCE) \cite{zhang2023dynamic} that proposed sensing local information dynamically. In contrast, our Local Context Fusion Module (LCFM) utilizes RoI-to-RoI self-attention to resolve spatial interactions without the need for manual surrounding searches. For that we need to model object interactions among the proposals generated earlier, by applying self-attention mechanism over the features generated by RoI. Let $\mathbf{R}=[\mathbf{r}_1,\dots,\mathbf{r}_N]^\top \in \mathbb{R}^{N\times d}$. Then we compute the following query, key, and value embeddings using learnable linear projections based on self-attention expessed as:
\begin{equation}
\mathbf{q}_i = \mathbf{W}_q \mathbf{r}_i,\quad
\mathbf{k}_j = \mathbf{W}_k \mathbf{r}_j,\quad
\mathbf{v}_j = \mathbf{W}_v \mathbf{r}_j,
\label{eq:qkv}
\end{equation}
where, $\mathbf{W}_q,\mathbf{W}_k,\mathbf{W}_v \in \mathbb{R}^{d_a \times d}$ and $d_a$ is the attention embedding dimension.
Then attention weight from RoI $i$ to RoI $j$ is computed as:
\begin{equation}
\alpha_{ij} =
\frac{\exp\left(\mathbf{q}_i^\top \mathbf{k}_j / \sqrt{d_a}\right)}
{\sum_{m=1}^{N}\exp\left(\mathbf{q}_i^\top \mathbf{k}_m / \sqrt{d_a}\right)}.
\label{eq:alpha}
\end{equation}
The local context feature for RoI $i$ is obtained by aggregating value embeddings:
\begin{equation}
\mathbf{c}_i^{\mathrm{loc}} = \sum_{j=1}^{N}\alpha_{ij}\mathbf{v}_j.
\label{eq:cloc}
\end{equation}
Finally, we form the locally enhanced RoI representation via residual fusion:
\begin{equation}
\tilde{\mathbf{r}}_i = \mathbf{r}_i + \mathbf{W}_{\mathrm{loc}}\,\mathbf{c}_i^{\mathrm{loc}},
\label{eq:rtilde}
\end{equation}
where, $\mathbf{W}_{\mathrm{loc}} \in \mathbb{R}^{d \times d_a}$. This module enables each RoI to incorporate information from other objects in the scene, improving robustness in crowded and occluded driving scenarios.

\subsection{Object-Centric Global Context Attention Modeling with Geometry Bias}
\label{sec:global}
To capture scene-level dependencies, we employ a global context fusion strategy inspired by the query-independent formulation presented in \cite{cao2019gcnet}, which allows efficient long-range modeling without the computational overhead of standard non-local blocks. This global prior is then integrated with local features presented earlier to resolve semantic ambiguities in complex urban environments \cite{zhang2023dynamic}.

Local context captures short-range interactions but does not provide a compact representation of the overall object configuration of the scene. We therefore, construct an \emph{object-centric global context vector} by aggregating RoI features rather than pixel-level feature maps in Global Context Attention Module (GCAM).

\subsubsection{Top-$K$ selection}
From the locally enhanced RoIs $\{\tilde{\mathbf{r}}_i\}_{i=1}^{N}$ in LCFM presented earlier, we select the top-$K$ proposals based on objectness or classification confidence, yielding:
\begin{equation}
\mathcal{S}=\{(\tilde{\mathbf{r}}_k,\mathbf{b}_k)\}_{k=1}^{K},\quad K \ll N,
\label{eq:topk}
\end{equation}
where $\mathbf{b}_k=(x_k,y_k,w_k,h_k)$ denotes the corresponding RoI box in image coordinates.

\subsubsection{Geometry-aware attention scoring}

We incorporate a lightweight geometry bias into the attention logits. Specifically, we define a normalized geometry encoding:
\begin{equation}
\phi(\mathbf{b}_k) =
\Big[\frac{x_k}{W},\,\frac{y_k}{H},\,\log\frac{w_k}{W},\,\log\frac{h_k}{H}\Big] \in \mathbb{R}^{4},
\label{eq:geom}
\end{equation}
here, $W,H$ are the image width and height. The geometry-aware attention logit for RoI $k$ is:
\begin{equation}
s_k =
\underbrace{\mathbf{u}^\top \sigma(\mathbf{W}_g \tilde{\mathbf{r}}_k)}_{\text{content}}
\;+\;
\underbrace{\mathbf{v}^\top \phi(\mathbf{b}_k)}_{\text{geometry bias}},
\label{eq:logit}
\end{equation}
where $\mathbf{W}_g \in \mathbb{R}^{d_g \times d}$, $\mathbf{u} \in \mathbb{R}^{d_g}$, $\mathbf{v} \in \mathbb{R}^{4}$ are learnable parameters, and $\sigma(\cdot)$ is a nonlinearity (e.g., ReLU). Then we normalize logits after applying Softmax.

The object-centric global context is computed as an attention-weighted sum of RoI features:
\begin{equation}
\mathbf{g}^{\mathrm{obj}} = \sum_{k=1}^{K}\beta_k \tilde{\mathbf{r}}_k \in \mathbb{R}^{d}.
\label{eq:gobj}
\end{equation}
The above Eq.~\eqref{eq:gobj} encourages the model to emphasize RoIs whose spatial layout is informative for driving scenes (e.g., traffic lights frequently appear in upper regions), while remaining lightweight and fully compatible with end-to-end training.

\subsubsection{Global context injection}

We broadcast $\mathbf{g}^{\mathrm{obj}}$ to all RoIs using a learnable projection as follows:
\begin{equation}
\hat{\mathbf{r}}_i = \tilde{\mathbf{r}}_i + \mathbf{W}_{\mathrm{glob}}\,\mathbf{g}^{\mathrm{obj}},
\label{eq:inject}
\end{equation}
where $\mathbf{W}_{\mathrm{glob}} \in \mathbb{R}^{d \times d}$.

\subsection{Context Feature Fusion and Detection Heads}
\label{sec:fusion}
We fuse appearance, local, and object-centric global context cues to form the final RoI representation. Using the original RoI feature $\mathbf{r}_i$, the local context feature $\mathbf{c}_i^{\mathrm{loc}}$ from Eq.~\eqref{eq:cloc}, and the global context vector $\mathbf{g}^{\mathrm{obj}}$ from Eq.~\eqref{eq:gobj}, we compute:
\begin{equation}
\mathbf{f}_i = \mathrm{MLP}\Big([\mathbf{r}_i \,\|\, \mathbf{c}_i^{\mathrm{loc}} \,\|\, \mathbf{g}^{\mathrm{obj}}]\Big),
\label{eq:fuse}
\end{equation}
where $\|\,$ denotes concatenation and $\mathrm{MLP}(\cdot)$ is a lightweight projection network. The fused feature $\mathbf{f}_i$ is fed into standard detection heads:
\begin{equation}
\hat{\mathbf{p}}_i = \mathrm{ClsHead}(\mathbf{f}_i),\qquad
\hat{\mathbf{b}}_i = \mathrm{RegHead}(\mathbf{f}_i),
\label{eq:heads}
\end{equation}
where $\hat{\mathbf{p}}_i$ denotes class probabilities and $\hat{\mathbf{b}}_i$ denotes predicted bounding boxes.

Finally, CCFF is trained end-to-end and the following loss function is calculated.

\begin{equation}
\mathcal{L} = \mathcal{L}_{\mathrm{cls}} + \lambda\mathcal{L}_{\mathrm{reg}},
\label{eq:ltotal}
\end{equation}
where $\mathcal{L}_{\mathrm{cls}}$ is the classification loss, $\mathcal{L}_{\mathrm{reg}}$ is the regression loss, and $\lambda$ is a hyperparameter used to balance the two terms.

Because the local attention module (Eqs.~\eqref{eq:qkv}--\eqref{eq:rtilde}) and geometry-aware object-centric attention pooling (Eqs.~\eqref{eq:topk}--\eqref{eq:inject}) are differentiable, gradients from Eq.~\eqref{eq:ltotal} propagate through the fusion stage (Eq.~\eqref{eq:fuse}) into both context modules and the backbone. The top-$K$ selection in Eq.~\eqref{eq:topk} is discrete; however, the attention pooling weights and subsequent fusion remain fully differentiable and train stably in practice.

\section{Experimental Results and Analysis}
\label{sec:experiments}

\subsection{Evaluation Metrics}
\label{ssec:subhead}
We evaluate our proposed model by considering two specialized metrics beyond standard object detection metrics. Our Co-occurring AP (CoAP) measures the precision of detections within contextual pairs, focusing on the model's ability to resolve individual objects through their spatial and semantic relationships \cite{lu2016visual}, and Category-level Consistency Strategy (CCS) \cite{Zhu2022COCM}  quantify the alignment between the predicted and ground-truth object co-occurrence distributions \cite{chen2017spatial}. This indicates that standard AP treats detection as independent events, while these metrics collectively demonstrate how effectively our Local Context Fusion Module (LCFM) and Global Context Attention Module (GCAM) capture the underlying structural logic of complex driving scenes, ensuring that high-confidence detections remain semantically and contextually enhanced as proposed in our research.

\subsection{Implementation Details}
\label{sec:impl}
Our proposed CCFF framework is implemented using two-stage detector framework- Detectron2 that works on Faster R-CNN based FPN backbone. Then the proposed modules- LCFM and GCAM are inserted into the RoI head, as mentioned earlier, and trained end-to-end using the standard detection losses, reported in Eq.~\eqref{eq:ltotal}. Unless otherwise stated, we adopt the default Detectron2 model training schedule and augmentations commonly used for these datasets. We set the number of selected proposals for global context $K=\{ \text{32 or 64} \}$ and keep the attention embedding dimension $d_a$ lightweight to limit overhead. 
We used Cityscapes \cite{cordts2016cityscapes} and BDD100K \cite{yu2020bdd100k} datasets with resolutions ($ 2048 \times 1024 $), and ($ 1280 \times 720 $), respectively. 
All models are trained using SGD with a fixed learning rate schedule. To ensure fair comparison among the proposed model variants, we considered the same backbone, optimizer settings, batch size, and categories. 
\begin{table}[t]
\centering
\caption{Comparative Analysis of the proposed approach on Cityscapes (all variants)}
\label{tab:cityscapes_all_variants}
\resizebox{\columnwidth}{!}{%
\begin{tabular}{l|ccc|ccc|cc}
\hline
\textbf{Experiment} & \textbf{AP} $\uparrow$ & \textbf{AP$_{50}$} $\uparrow$ & \textbf{AP$_{75}$} $\uparrow$ & \textbf{AP$_{S}$} & \textbf{AP$_{M}$} & \textbf{AP$_{L}$} & \textbf{CoAP} & \textbf{CCS } \\ \hline
Baseline & \textbf{36.44} & \textbf{60.59} & \textbf{36.11} & \textbf{12.03} & 34.68 & \textbf{60.89} & 0.386 & 0.972 \\
Local & 35.34 & 59.12 & 34.12 & 11.08 & 34.88 & 59.72 & 0.382 & 0.972 \\
Global & 35.41 & 59.25 & 34.22 & 11.12 & 34.92 & 59.81 & 0.383 & 0.972 \\
\textbf{Ours} & 35.47 & 59.35 & 34.39 & 11.19 & \textbf{34.99} & 59.87 & \textbf{0.389} & \textbf{0.973} \\
Ours\_geom & 35.51 & 59.42 & 34.45 & 11.22 & 35.05 & 59.92 & 0.385 & 0.972 \\ \hline
\end{tabular}%
}
\begin{flushleft}
\small
\end{flushleft}
\end{table}

\subsection{Quantitative Analysis on Cityscapes}
We reported the quantitative performance analysis of our \textit{CCFF} variants on the Cityscapes validation set in Table~\ref{tab:cityscapes_all_variants}. This shows that the baseline achieves overall $AP$ performance gain, our \textit{Ours} variant demonstrates better capabilities in relational reasoning and scale-specific detection. We report scale-specific precision and our model demonstrates strong retrieval capabilities, particularly for medium-scale entities. Similarly, the proposed CCFF framework not only localizes objects accurately but also maintains a high capture rate for critical road factors.

\textbf{Relational Consistency and Co-occurrence:} 
As reported in Row 4 in Table ~\ref{tab:cityscapes_all_variants}, \textbf{Ours} variant achieves a \textbf{CoAP}  of \textbf{0.389} and a \textbf{CCS} of \textbf{0.973}, outperforming all other configurations, including the baseline. This improvement indicates that our dual-stream context fusion successfully captures semantic dependencies between co-occurring objects. Additionally, the higher CCS score indicates that our model maintains higher spatial rank correlation, which is important for scene understanding in autonomous driving.

\textbf{Scale-Specific Performance:} 
In \textbf{(AP$_{M}$)} column of Table  ~\ref{tab:cityscapes_all_variants}, \textit{Ours} variant reaches \textbf{34.99\%} surpassing the baseline's 34.68\%. This suggests that our object-centric fusion logic is particularly effective for medium-range interactions, where objects are close enough to provide mutual context. From this, we observe that feature enhancement from neighboring instances is required to resolve ambiguities.

\textbf{Ablation of Fusion Variants:} 
Comparing the \textit{Local} and \textit{Global} variants reveals that global scene context ($AP$ 35.41\%) provides a slight edge over purely local neighborhood features ($AP$ 35.34\%). However, fusion of both in \textit{Ours} model provides the most balanced precision-context trade-off. Finally, the \textit{Ours\_geom} variant gives the highest raw $AP$ (35.51\%) among our proposed heads, ensuring that geometric priors such as distance and orientation further refine bounding box localization.

\subsection{Quantitative Analysis on BDD100K}
Our Table~\ref{tab:bdd100k_final} shows the performance of the proposed variants on the BDD100K dataset. Here, our object-centric approach demonstrates superior efficacy in capturing complex urban dependencies and resolving small-object localization challenges inherent in the BDD100K benchmark.

\textbf{Relational Integrity and Spatial Consistency:}
The proposed \textit{Ours} variant (shown in bold in Row 4) achieves \textbf{CoAP} of \textbf{0.488} and \textbf{CCS} of \textbf{0.969}, significantly outperforming the baseline. This lead indicates that our dual-stream fusion head is uniquely robust at maintaining spatial ranking and contextual precision, even in the diverse lighting and weather conditions (main characteristic of BDD100K dataset).

\textbf{Small Object Detection Gains:}
A significant performance improvement is observed in the small object category of column ($AP_S$) in \textit{Ours} method is \textbf{14.73\%}. Here, by leveraging relational information from high-confidence neighboring object proposal embeddings, our model effectively improves the feature representation of distant or ambiguous objects, and as a result, our final overall \textbf{AP} is \textbf{32.95\%}, establishing a competitive baseline for object-centric autonomous driving research.

\textbf{Geometric Refinement:}
By adding relative spatial priors in the \textit{Ours\_geom} variant (reported in Row 5 in Table ~\ref{tab:bdd100k_final}), further optimization of localization is achieved, yielding an overall \textbf{AP} of \textbf{32.21\%}. These results show that explicit modeling of distance and orientation in highly competitive environments provides a crucial inductive bias for original scene understanding.

\begin{table}[t]
\centering
\caption{Object Detection and Category-level Consistency Strategy Results on BDD100K (all variants)}
\label{tab:bdd100k_final}
\resizebox{\columnwidth}{!}{%
\begin{tabular}{l|ccc|ccc|cc}
\hline
\textbf{Variant} & \textbf{AP} $\uparrow$ & \textbf{AP$_{50}$} $\uparrow$ & \textbf{AP$_{75}$} $\uparrow$ & \textbf{AP$_{S}$} & \textbf{AP$_{M}$} & \textbf{AP$_{L}$} & \textbf{CoAP} & \textbf{CCS} $\uparrow$ \\ \hline
Baseline & 31.02 & 54.21 & 30.50 & 12.12 & 32.08 & 52.12 & 0.408 & 0.952 \\
Local & 31.21 & 30.08 & 30.81 & 12.44 & 33.11 & 52.42 & 0.410 & 0.954 \\
Global & 31.41 & 54.81 & 31.09 & 12.06 & 33.42 & 52.81 & 0.415 & 0.955 \\
\textbf{Ours} & \textbf{32.95} & \textbf{56.89} & \textbf{32.92} & \textbf{14.73} & \textbf{35.88} & \textbf{54.94} & \textbf{0.488} & \textbf{0.969} \\
Ours\_geom & 32.21 & 56.28 & 32.51 & 14.34 & 35.51 & 54.81 & 0.431 & 0.965 \\ \hline
\end{tabular}%
}
\end{table}

\textbf{Contextual Constraints and Transformer Baselines:} While global query-based transformers implicitly model multi-scale features, their lack of explicit localized geometric constraints often makes tiny, low-frequency, or heavily occluded categories difficult to capture under strict edge latency scenarios. In contrast, our dual-stream CCFF head explicitly adds geometric inductive biases into the regional level Eq.~\eqref{eq:logit}. This design allows our model to achieve a significant performance enhancement in small-scale category ($AP_S$: 14.73\% on BDD100K) over standard two-stage baselines. Crucially, as detailed in our efficiency benchmarks (Sec. 4.5), this structural precision is achieved with an almost negligible latency penalty of $\approx$ 0.2 FPS. This indicates that dedicated context-centric feature fusion yields superior performance-to-latency trade-offs for real-world driving environments over generic, unconstrained pixel-level transformer baselines.

\begin{figure*}[t]
  \centering
  \includegraphics[width=1.0\linewidth]{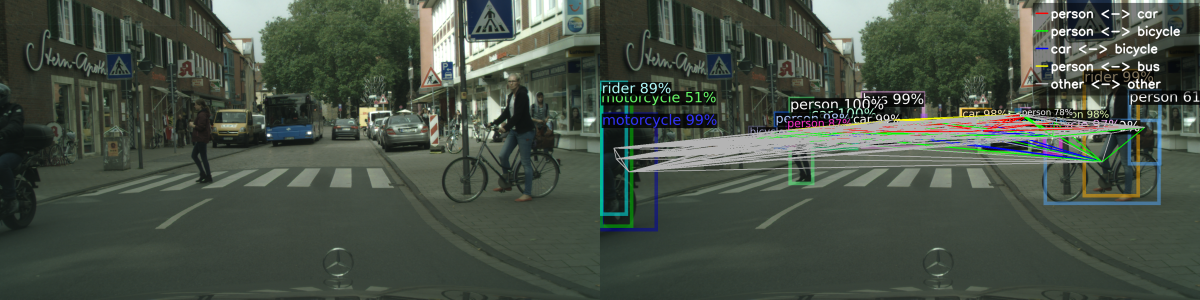}
  \caption{\textbf{Qualitative visualization of semantic co-occurrence links during inference on Cityscapes.} The left panel displays the raw input scene. The right panel illustrates finalized model predictions with our explicit relational logic links. Line colors distinguish discrete category configurations (e.g., red for \textit{person} $\leftrightarrow$ \textit{car}, green for \textit{person} $\leftrightarrow$ \textit{bicycle}). Line widths reflect attention confidence, demonstrating how the model utilizes highly visible anchors to stabilize the classification of ambiguous background elements.}

  \label{fig:cooccurrence_viz_0}
\end{figure*}

We evaluate the effectiveness of our proposed \textit{CCFF} framework by visualizing the semantic relationships captured during inference on the Cityscapes dataset. As illustrated in Fig.~\ref{fig:cooccurrence_viz_0}, we present the original input image (left) alongside the corresponding detections and co-occurrence edges generated by our framework (right). Our model explicitly maps co-occurrences of objects through a network of colored semantic edges, such as \textit{person} $\leftrightarrow$ \textit{car} and \textit{person} $\leftrightarrow$ \textit{bicycle}. Using CCS, the framework is able to reason about the structural layout of the scene, achieving high-confidence detections even for challenging ``tail'' categories (e.g., \textit{rider} 89\%, \textit{motorcycle} 99\%) in cluttered urban environments. The dense network of edges demonstrates how our local-global feature fusion leverages co-occurrence data to resolve ambiguities and provide a more holistic understanding of the scene compared to standard localized detection pipelines.

\begin{figure*}[t]
  \centering
  \includegraphics[width=1.0\linewidth]{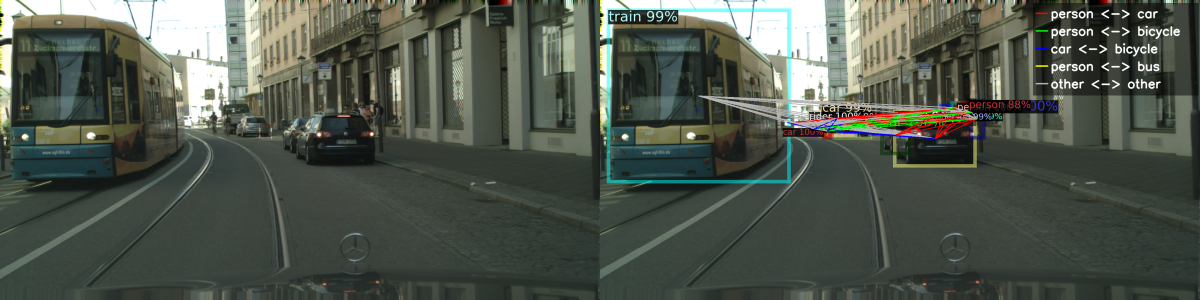}
  \caption{\textbf{Qualitative results illustrating scale robustness in heterogeneous urban driving environments.} The left panel represents the raw street environment. The right panel displays the corresponding CCFF inference result. Our model simultaneously maps the co-occurrences by utilizing a highly confident macro landmark (e.g., the 99\% confidence \textit{train} extraction) as a structural anchor. The co-occurrence links (e.g., yellow line represents \textit{person} $\leftrightarrow$ \textit{bus} ) act as a contextual representation to confidently identify and recover distant or heavily occluded background instances.}

  \label{fig:tram_cooccurrence}
\end{figure*}

Similarly, Fig.~\ref{fig:tram_cooccurrence}, visualizes the semantic relationship captured during inference on the Cityscapes dataset. Here, our proposed framework exhibits high robustness in complex street scenes containing diverse object scales. In this scenario, the \textit{CCFF} accurately detects a large-scale tram (\textbf{train 99\%}) while simultaneously identifying several smaller cars and pedestrians in the middle-to-background regions. By using CCS, our model generates a dense network of edges that link these heterogeneous objects. Specifically, the presence of the highly-confident train detection acts as a contextual anchor, where the resulting semantic edges—including the \textit{person} $\leftrightarrow$ \textit{car} (red) and other co-occurrences (white)—provide global relational cues that reinforce the detection of more distant or partially occluded instances. This explicit modeling of spatial and semantic proximity ensures that the final predictions remain consistent with the structural layout of the urban environment, achieving a more holistic scene interpretation than traditional, localized detection methods.

\subsection{Efficiency Analysis}
\label{efficiencyanalysis}
An efficiency analysis is reported in Table \ref{tab:efficiency_single_column} reveals that our \textit{CtxFusion} module adds 5.38M parameters, representing a manageable 13\% increase in model size over the baseline. In terms of latency, the overhead remains minimal across varying resolutions: BDD100K ($1280 \times 720$) incurs an additional 7.7\,ms, while the high-resolution Cityscapes ($2048 \times 1024$) sees a 16.5\,ms increase. Consequently, our method maintains real-time viability with only a marginal FPS reduction ($\approx 0.2$), demonstrating that our object-centric improvements do not sacrifice practical deployment speed.

\begin{table}[h] 
\centering 
\caption{Model Complexity and Inference Efficiency. All models use ResNet-50 + FPN backbone and are evaluated on a single GPU.} 
\label{tab:efficiency_single_column} 
\resizebox{\columnwidth}{!}{%
\begin{tabular}{l|lccc} 
\hline 
\textbf{Dataset} & \textbf{Variant} & \textbf{Params (M)} & \textbf{Lat. (ms)} & \textbf{FPS} \\ \hline 
Cityscapes & Baseline & 41.33 & 308.04 & 3.25 \\ 
 & \textbf{Ours} & \textbf{46.71} & \textbf{324.57} & \textbf{3.08} \\ \hline 
BDD100K & Baseline & 41.33 & 160.53 & 6.23 \\ 
 & \textbf{Ours} & \textbf{46.71} & \textbf{168.24} & \textbf{5.94} \\ \hline 
\textit{Overhead ($\Delta$)} & \textit{Added} & \textit{+5.38} & \textit{+7.7--16.5} & \textit{$\approx$-0.2} \\ \hline 
\end{tabular}%
} 
\end{table}

\subsection{Ablation Study}
\label{sec:ablation}

We conduct ablation study to evaluate the performance of our \textit{CtxFusion} components, as summarized in Table~\ref{tab:ablation_study}. The results indicate that while individual \textbf{Local} or \textbf{Global} modules establish stable relational baselines, their combination is required to effectively capture complex urban dependencies.

\textbf{Effect of Dual-Stream Fusion:} By analyzing the results reported in Table~\ref{tab:ablation_study} we can understand that integrating local neighborhood interactions with global scene context (Row 3) provides the most significant performance leap. In the diverse BDD100K dataset, this dual-stream approach achieves a peak AP of 32.95 and a substantial boost in contextual precision (CoAP: 0.488). This suggests that modeling the relationship between an object and its immediate neighbors, while simultaneously accounting for the overall scene configuration, is essential for resolving ambiguities in complex driving environments.

\textbf{Role of Geometric Priors:} The addition of geometric embeddings $\phi(\mathbf{b}_{k})$ in the \textbf{Ours\_geom} variant (Row 4) acts as a structural stabilizer. While the content-only fusion in Row 3 provides higher raw AP on BDD100K, the geometric variant maintains high reliability across datasets. By anchoring features to physical coordinates and scales, the model moves toward a structural reasoning approach that effectively filters background noise and resolves position ambiguities for distant objects. 

\textbf{Co-occurrence Supervision:} We observe that the model's ability to maintain high \textit{CoAP} and \textit{CCS} scores depend on the joint optimization of the local and global modules. Even without a separate supervision term, the dual-stream architecture implicitly learns semantic co-occurrence priors. 

\begin{table}[h]
\centering
\caption{Ablation study of context components on Cityscapes and BDD100K validation sets.}
\label{tab:ablation_study}
\resizebox{\columnwidth}{!}{%
\begin{tabular}{ccc|ccc|ccc}
\hline
\multicolumn{3}{c|}{\textbf{Components}} & \multicolumn{3}{c|}{\textbf{Cityscapes}} & \multicolumn{3}{c}{\textbf{BDD100K}} \\
\textbf{Loc.} & \textbf{Glob.} & \textbf{Geom.} & \textbf{AP} & \textbf{CoAP} & \textbf{CCS} & \textbf{AP} & \textbf{CoAP} & \textbf{CCS} \\ \hline
\multicolumn{3}{c|}{Baseline} & \textbf{36.44} & 0.386 & 0.972 & 31.02 & 0.408 & 0.952 \\ \hline
\checkmark & $\times$ & $\times$ & 35.34 & 0.382 & 0.972 & 31.21 & 0.410 & 0.954 \\
$\times$ & \checkmark & $\times$ & 35.41 & 0.383 & 0.972 & 31.41 & 0.415 & 0.955 \\
\checkmark & \checkmark & $\times$ & 35.47 & \textbf{0.389} & \textbf{0.973} & \textbf{32.95} & \textbf{0.488} & \textbf{0.969} \\
\checkmark & \checkmark & \checkmark & 35.51 & 0.385 & 0.972 & 32.21 & 0.431 & 0.965 \\ \hline
\end{tabular}%
}
\end{table}
\section{Conclusion}
\label{sec:conclusion}
In this paper, we introduce Context-Centric Feature Fusion (CCFF), a framework that improves autonomous driving perception by leveraging explicit relational reasoning. Rather than treating objects in isolation, CCFF uses two key components: the Local Context Fusion Module (LCFM) to capture spatial relationships between regions, and the Global Context Attention Module (GCAM) to model how objects typically co-occur. Together, these modules shift the focus from independent detection to holistic scene understanding. Our evaluations on Cityscapes and BDD100K show significant gains, especially for small or rare objects that are often misidentified. Furthermore, a higher CCS  confirms that our model successfully learns the underlying structural layout of urban environments.

{
 \small
 \bibliographystyle{ieeenat_fullname}

\begin{thebibliography}{16}
\providecommand{\natexlab}[1]{#1}
\providecommand{\url}[1]{\texttt{#1}}
\expandafter\ifx\csname urlstyle\endcsname\relax
  \providecommand{\doi}[1]{doi: #1}\else
  \providecommand{\doi}{doi: \begingroup \urlstyle{rm}\Url}\fi

\bibitem[Bell et~al.(2016)Bell, Zitnick, Bala, and Girshick]{Bell2016ION}
Sean Bell, C.~Lawrence Zitnick, Kavita Bala, and Ross Girshick.
\newblock Inside-outside net: Detecting objects in context with skip pooling and recurrent neural networks.
\newblock In \emph{Proceedings of the IEEE Conference on Computer Vision and Pattern Recognition (CVPR)}, pages 2874--2883, 2016.

\bibitem[Cao et~al.(2019)Cao, Xu, Lin, Wei, and Hu]{cao2019gcnet}
Yue Cao, Jiarui Xu, Stephen Lin, Fangyun Wei, and Han Hu.
\newblock {GCNet}: Non-local networks meet squeeze-excitation networks and beyond.
\newblock In \emph{Proceedings of the IEEE/CVF International Conference on Computer Vision (ICCV) Workshops}, pages 1971--1980, 2019.

\bibitem[Carion et~al.(2020)Carion, Massa, Synnaeve, Usunier, Kirillov, and Zagoruyko]{carion2020end}
Nicolas Carion, Francisco Massa, Gabriel Synnaeve, Nicolas Usunier, Alexander Kirillov, and Sergey Zagoruyko.
\newblock End-to-end object detection with transformers.
\newblock In \emph{Proceedings of the European Conference on Computer Vision (ECCV)}, pages 213--229, 2020.

\bibitem[Chen and Gupta(2017)]{chen2017spatial}
Xinlei Chen and Abhinav Gupta.
\newblock Spatial memory for context reasoning in object detection.
\newblock In \emph{Proceedings of the IEEE International Conference on Computer Vision (ICCV)}, pages 4087--4096, 2017.

\bibitem[Cordts et~al.(2016)Cordts, Omran, Ramos, Scharw{\"a}chter, Enzweiler, Benenson, Franke, Roth, and Schiele]{cordts2016cityscapes}
Marius Cordts, Mohamed Omran, Sebastian Ramos, Timo Scharw{\"a}chter, Markus Enzweiler, Rodrigo Benenson, Uwe Franke, Stefan Roth, and Bernt Schiele.
\newblock The {Cityscapes} dataset for semantic urban scene understanding.
\newblock In \emph{Proceedings of the IEEE Conference on Computer Vision and Pattern Recognition (CVPR)}, pages 3213--3223, 2016.

\bibitem[Guo et~al.(2025)Guo, Jiang, Chen, Wang, Sha, and Chen]{Guo2025}
Xuyao Guo, Feng Jiang, Quanzhen Chen, Yuxuan Wang, Kaiyue Sha, and Jing Chen.
\newblock Deep learning-enhanced environment perception for autonomous driving: {MDNet} with {CSP-DarkNet53}.
\newblock \emph{Pattern Recognition}, 160:\penalty0 111174, 2025.

\bibitem[Hu et~al.(2018)Hu, Gu, Zhang, Dai, and Wei]{hu2018relationnet}
Han Hu, Jiayuan Gu, Zheng Zhang, Jifeng Dai, and Yichen Wei.
\newblock Relation networks for object detection.
\newblock In \emph{Proceedings of the IEEE Conference on Computer Vision and Pattern Recognition (CVPR)}, pages 3588--3597, 2018.

\bibitem[Hua et~al.(2025)Hua, Wu, Hao, Xia, and Li]{hua2025elft}
Guoguang Hua, Fangfang Wu, Guangzhao Hao, Chenbo Xia, and Li Li.
\newblock {ELFT}: Efficient local-global fusion transformer for small object detection.
\newblock \emph{PLoS ONE}, 20\penalty0 (9):\penalty0 e0332714, 2025.

\bibitem[Li et~al.(2025)Li, Cheang, Yu, Tang, Du, and Cheng]{Li2025}
Jiayao Li, Chak~Fong Cheang, Xiaoyuan Yu, Suigu Tang, Zhaolong Du, and Qianxiang Cheng.
\newblock A segmentation network for enhancing autonomous driving scene understanding using skip connection and adaptive weighting.
\newblock \emph{Scientific Reports}, 15\penalty0 (1):\penalty0 36692, 2025.

\bibitem[Lu et~al.(2016)Lu, Krishna, Bernstein, and Fei-Fei]{lu2016visual}
Cewu Lu, Ranjay Krishna, Michael Bernstein, and Li Fei-Fei.
\newblock Visual relationship detection with language priors.
\newblock In \emph{Proceedings of the IEEE Conference on Computer Vision and Pattern Recognition (CVPR)}, pages 852--860, 2016.

\bibitem[Mottaghi et~al.(2014)Mottaghi, Chen, Liu, Cho, Lee, Fidler, Urtasun, and Yuille]{mottaghi2014role}
Roozbeh Mottaghi, Xianjie Chen, Xiaobai Liu, Nam-Gyu Cho, Seong-Whan Lee, Sanja Fidler, Raquel Urtasun, and Alan Yuille.
\newblock The role of context for object detection and semantic segmentation in the wild.
\newblock In \emph{Proceedings of the IEEE Conference on Computer Vision and Pattern Recognition (CVPR)}, pages 891--898, 2014.

\bibitem[Wang et~al.(2018)Wang, Girshick, Gupta, and He]{wang2018nonlocal}
Xiaolong Wang, Ross Girshick, Abhinav Gupta, and Kaiming He.
\newblock Non-local neural networks.
\newblock In \emph{Proceedings of the IEEE Conference on Computer Vision and Pattern Recognition (CVPR)}, pages 7794--7803, 2018.

\bibitem[Yu et~al.(2020)Yu, Chen, Wang, Xian, Chen, Liu, Madhavan, and Darrell]{yu2020bdd100k}
Fisher Yu, Haofeng Chen, Xin Wang, Wenqi Xian, Yingying Chen, Fangchen Liu, Vashisht Madhavan, and Trevor Darrell.
\newblock {BDD100K}: A diverse driving dataset for heterogeneous multitask learning.
\newblock In \emph{Proceedings of the IEEE/CVF Conference on Computer Vision and Pattern Recognition (CVPR)}, pages 2636--2645, 2020.

\bibitem[Zhang et~al.(2023)Zhang, Gong, Sun, Wu, and Yang]{zhang2023dynamic}
Ziji Zhang, Ping Gong, Haotian Sun, Pingping Wu, and Xuanyuan Yang.
\newblock Dynamic local and global context exploration for small object detection.
\newblock In \emph{Proceedings of the IEEE International Conference on Acoustics, Speech and Signal Processing (ICASSP)}, pages 1--5, 2023.

\bibitem[Zhu et~al.(2022)Zhu, Tian, Zhou, Bai, and Song]{Zhu2022COCM}
Siyu Zhu, Yingjie Tian, Fenfen Zhou, Kunlong Bai, and Xiaoyu Song.
\newblock {COCM}: Co-occurrence-based consistency matching in domain-adaptive segmentation.
\newblock \emph{Mathematics}, 10\penalty0 (23):\penalty0 4468, 2022.

\bibitem[Zhu et~al.(2021)Zhu, Su, Lu, Li, Wang, and Dai]{zhu2021deformable}
Xizhou Zhu, Weijie Su, Lewei Lu, Bin Li, Xiaogang Wang, and Jifeng Dai.
\newblock Deformable {DETR}: Deformable transformers for end-to-end object detection.
\newblock In \emph{International Conference on Learning Representations (ICLR)}, 2021.

\end{thebibliography}

}

\end{document}